%% file: main.tex
\documentclass[twoside,leqno,twocolumn]{article}

\usepackage[letterpaper]{geometry}

\usepackage{amsfonts,amsmath,amssymb}
\usepackage{ltexpprt}
\usepackage{hyperref}
\usepackage{booktabs}
\usepackage{graphicx}
\usepackage{wasysym}
\usepackage{float}
\usepackage{xcolor}
\usepackage{arydshln}
\usepackage{subcaption}
\usepackage{multirow}
\usepackage{enumitem}
\usepackage[normalem]{ulem}
\usepackage{algorithm}
\usepackage{algpseudocode}

\usepackage[symbol]{footmisc}

\begin{document}

\newcommand\relatedversion{}

\title{
    \Large Task Aware Modulation using Representation Learning: An Approach for Few Shot Learning in Environmental Systems
}
\author{
Arvind Renganathan\thanks{University of Minnesota. \{renga016, ghosh128, khand035, kumar001\}@umn.edu} ${\footnotemark[2]}$ \and
Rahul Ghosh${\footnotemark[1]}$ ${\footnotemark[2]}$ \and
Ankush Khandelwal${\footnotemark[1]}$ \and
Vipin Kumar${\footnotemark[1]}$
}

\date{}

\maketitle

$\footnotetext[2]{Equal contribution}$


\fancyfoot[R]{\thepage}





\begin{abstract}
\small\baselineskip=9pt We introduce TAM-RL (Task Aware Modulation using Representation Learning), a novel multimodal meta-learning framework for few-shot learning in heterogeneous systems, designed for science and engineering problems where entities share a common underlying forward model but exhibit heterogeneity due to entity-specific characteristics. TAM-RL leverages an amortized training process with a modulation network and a base network to learn task-specific modulation parameters, enabling efficient adaptation to new tasks with limited data. We evaluate TAM-RL on two real-world environmental datasets: Gross Primary Product (GPP) prediction and streamflow forecasting, demonstrating significant improvements over existing meta-learning methods. On the FLUXNET dataset, TAM-RL improves RMSE by 18.9\% over MMAML with just one month of few-shot data, while for streamflow prediction, it achieves an 8.21\% improvement with one year of data. Synthetic data experiments further validate TAM-RL's superior performance in heterogeneous task distributions, outperforming the baselines in the most heterogeneous setting. Notably, TAM-RL offers substantial computational efficiency, with at least 3x faster training times compared to gradient-based meta-learning approaches while being much simpler to train due to reduced complexity. Ablation studies highlight the importance of pretraining and adaptation mechanisms in TAM-RL's performance. 

\end{abstract}

\input{./1_introduction}

\input{./2_relatedWork}

\input{./3_problemFormulation}

\input{./4_method}

\input{./5_experiments}

\input{./6_results}

\input{./7_conclusion}

\bibliographystyle{plain}
\bibliography{main}

\clearpage
\appendix
    

\newpage
\input{./Appendix}

\end{document}

%% file: 1_introduction.tex
\vspace{-0.1in}
\section{Introduction}
\label{sec:introduction}

\begin{figure}
    \centering
    \includegraphics[width=0.9\columnwidth]{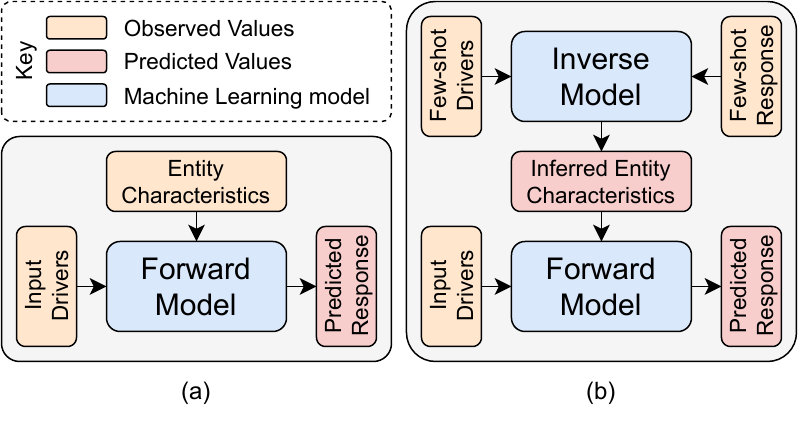}
    \vspace{-0.2in}
    \caption{\footnotesize \textit{(a) Forward modeling for an entity/physical system with known characteristics. (b) TAM-RL caricature}}
    \label{fig:forward_and_inverse}
    \vspace{-0.3in}
\end{figure}

Personalized prediction in entities is a crucial aspect of many real-world scenarios, where the goal is to forecast the response of an entity/task based on its drivers. As an illustrative example, consider Gross Primary Product (GPP) prediction across the globe, with grid cells representing entities, GPP as the response, and factors such as weather conditions (e.g., temperature, rainfall) and Leaf Area Index (LAI) as drivers. The relationship between drivers and response is often heterogeneous, influenced by entity characteristics. Here, we term this type of heterogeneity as entity characteristics-induced heterogeneity. In the GPP example, different grid cells, characterized by their unique eco-regions and climate zones, can exhibit vastly different GPP responses to identical weather conditions and LAI values ~\cite{lin2021improved}. Such scenarios are prevalent in environmental applications and also extend to other domains. For instance, in healthcare, individuals (entities) may display varying heart rates (response) for the same level of physical activity (driver), depending on their physical fitness (entity characteristic). These examples underscore the importance of considering entity-specific attributes when modeling the driver-response relationship, as shown in Figure~\ref{fig:forward_and_inverse}(a).

Process-based forward models (PBMs) are widely used in environmental applications to simulate complex physical phenomena. These models incorporate numerous parameters to capture the heterogeneity in driver-response relationships across different entities. PBMs are typically calibrated through an inverse process, where parameter values are estimated for each entity by running the forward model with various parameter combinations and identifying the set that best fits the available driver-response observations. While this calibration method works well for well-observed entities, it faces significant limitations when applied to sparsely observed entities. A notable issue is equifinality, where multiple parameter combinations representing distinct underlying realities produce equally acceptable results, leading to ambiguity in model interpretation~\cite{gupta2014debates}. Furthermore, a critical shortcoming of the PBM framework is its inability to leverage observations from well-observed entities to enhance predictions for sparsely observed or unobserved entities.

\begin{figure}
    \centering
    \includegraphics[width=\columnwidth]{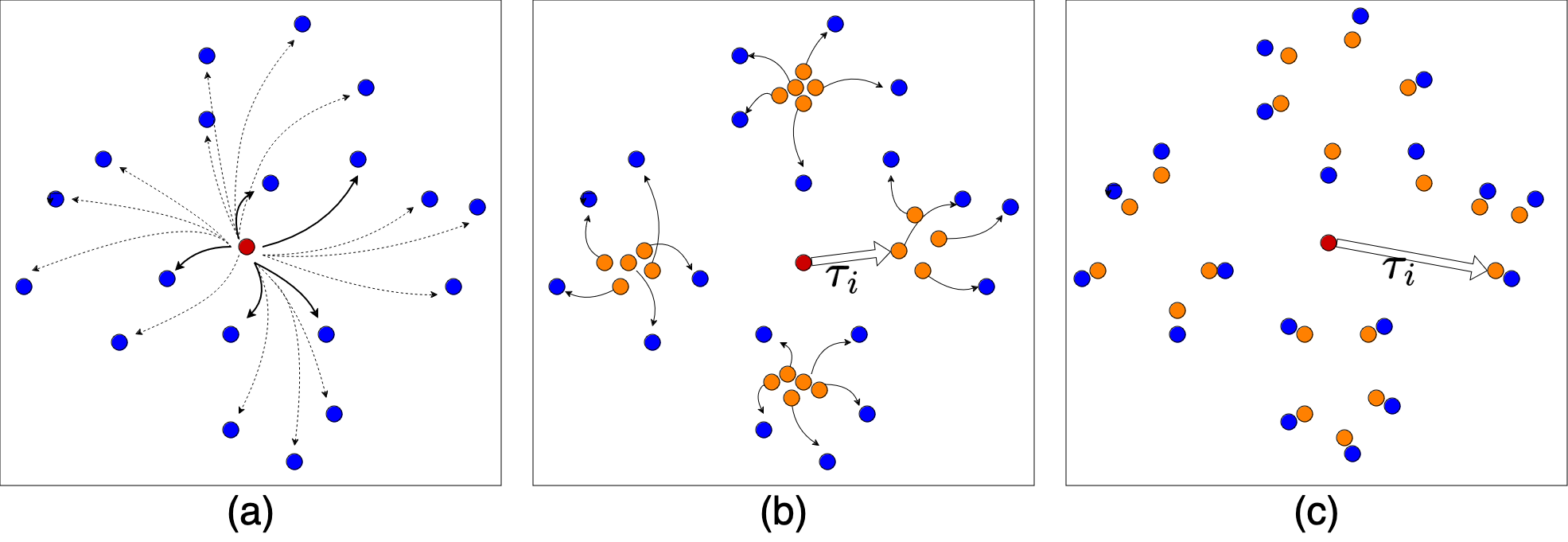}
    \vspace{-0.3in}
    \caption{\footnotesize \textit{Comparison of MAML, MMAML, and TAM-RL a) \textbf{MAML} computes single meta-initialization (red dot), which is adapted(curved arrows) to tasks(blue dots). MAML fails to adapt to dissimilar tasks (outer blue dots) using few shots of data b) \textbf{MMAML}~\cite{vuorio2019multimodal} uses task-specific modulation parameters to modulate (straight arrow) meta-initialization (red dot) and get task specific initialization (orange dot) which are then fine-tuned to tasks(blue dots). c) \textbf{TAM-RL} uses task-specific modulation parameters to modulate (straight arrow) shared initialization (red dot) to directly compute meta initialization (orange dots) for given tasks(blue dots).}}
    \label{fig:meta-learning}
    \vspace{-0.3in}
\end{figure}

Recently, machine learning (ML) models have shown increasing success in addressing the well-known limitations~\cite{jia2019physics,willard2022integrating} of PBMs. In particular, several works have already demonstrated the advantages of constructing an ML model that utilizes driver and response data from a set of well-observed entities (training set) and using them to make predictions on entities that are not present in the training set in domains such as engineering~\cite{kim2022predictive}, healthcare~\cite{logan2022patient}, and environmental sciences~\cite{kratzert2019towards}. Typically, these “zero-shot” approaches require entity characteristics~\cite{wang2019survey} to transfer information from entities in the training set to those in the test set ~\cite{kratzert2019toward}.

Entity characteristics are often unavailable or highly uncertain in many scenarios~\cite{beven2020deep}. Consequently, there is a need for learning methods that can generate personalized predictions for each entity based solely on a limited amount of (driver, response) data, without relying on prior knowledge of entity or task characteristics. Meta-learning~\cite{hospedales2021meta} is a popular few-shot learning framework that enables models to quickly learn new tasks in a few-shot setting by leveraging task similarities. Model agnostic meta-learning (MAML)~\cite{finn2017model} is one such approach that enables rapid adaptation to new tasks with limited training data, by performing a small number of gradient updates on meta-initialized parameters of a base network (Fig~\ref{fig:meta-learning}a). However, finding a single set of meta-initialized parameters for the base network that closely aligns with all task parameters is not possible when the task distribution is multi-modal (see Fig~\ref{fig:meta-learning}a). Several methods have been proposed to address this challenge~\cite{vuorio2019multimodal,arango2021multimodal,oreshkin2018tadam}. Typically, these methods involve first computing entity/task-specific embeddings from a task modulation network, which includes a task encoder and a modulation parameter generator. Given these entity embeddings, the meta-initialized parameters for the base prediction network are modulated to bring them closer to the entity/task groups in the multi-modal task distribution, see Figure~\ref{fig:meta-learning}(b). Similarly to MAML, both the task modulation and base networks are meta-trained using a few steps of gradient-based adaptation.

This paper presents Task Aware Modulation using Representation Learning (TAM-RL) framework, which reformulates knowledge transfer from well-observed to few-shot entities/tasks in environmental systems as two distinct processes: (1) inferring task-specific parameters from few-shot driver-response data, and (2) constructing a forward model parameterized by both task-specific and global parameters. TAM-RL achieves this through a unified training approach across all tasks/entities in the training set for the DL-based \textit{forward} and \textit{inverse models}, as illustrated in Figure~\ref{fig:forward_and_inverse}. TAM-RL uses the \textit{inverse model} to predict entity/task characteristics from few-shots of (driver, response) data, which are then used to infer task-specific modulation parameters via a \textit{parameter generator}, see Figure~\ref{fig:meta-learning}(c). Together, this \textit{inverse model} and \textit{parameter generator} are called the \textit{modulation network}. Concurrently, the \textit{forward model} is parameterized by these modulated task-specific meta-initialized parameters and shared global parameters for all entities/tasks. Hence, TAM-RL can be considered as an amortization-based meta-learning approach. This approach effectively creates an emulator for physics-based models with global parameters analogous to traditional PBMs. Once trained, the \textit{forward model} can predict future responses based on input drivers. 

We demonstrate the usefulness of TAM-RL by performing extensive evaluation on two publicly available real-world environmental datasets: a) predicting Gross Primary Product (GPP) in FLUXNET~\cite{pastorello2020fluxnet2015} a global dataset and b) forecasting streamflow in CARAVAN-GB~\cite{kratzert2022caravan} a national scale dataset. Specifically, we show that TAM-RL performs significantly better than several representative meta-learning methods. We additionally present an empirical evaluation via synthetic data to explore the impact of heterogeneity amongst the entities on the relative performance of TAM-RL. Further, we provide an extensive ablation study by evaluating each component of the proposed TAM-RL framework. Our main contributions are listed below:

$\bullet$ We present TAM-RL as a framework for building predictive models in a few-shot setting, for a set of entities for which heterogeneity is induced by unknown entity-specific characteristics. This framework involves joint training of forward and inverse models, where the inverse model is used to infer entity-specific characteristics, which are then used in the forward model to make personalized predictions.

$\bullet$ Extensive evaluation shows that TAM-RL outperforms the closest baseline by 10.3\% and 6.7\% in FLUXNET and CARAVAN-GB, respectively. Compared to baselines, the framework achieves its performance while being at least 45.26\% and 66.7\% more efficient in memory and runtime.

$\bullet$ We evaluate each component of TAM-RL through an ablation study 

$\bullet$ Finally, we also explore the impact of heterogeneity amongst the entities on TAM-RL's performance through empirical synthetic data evaluation.



%% file: 2_relatedWork.tex
\section{Related Work}
\label{sec:relatedWork}

Most recent successful approaches for few-shot learning use meta-learning~\cite{hospedales2021meta,huisman2021survey} to enable models to learn quickly from limited examples and adapt to new tasks efficiently. We categorize existing methods based on the following three criteria. For the first criterion, meta-learning methods can be categorized based on the number of parameters adapted for each task. While adapting more parameters increases flexibility and expressiveness, it risks overfitting and reduces data efficiency. Secondly, meta-learning methods can be categorized based on the choice of adaptation approach, ranging from amortized-based to gradient-based. Progressing from amortized-based approaches \cite{oreshkin2018tadam,abdollahzadeh2021revisit,requeima2019fast,patacchiola2020bayesian,sung2018learning} to gradient-based approaches\cite{finn2017model,antoniou2018train,jiang2018learning,baik2021meta,zintgraf2019fast} brings more flexibility in the adaptation approach but comes with higher computational expenses and a propensity for overfitting, leading to increased complexity and instability. Finally, meta-learning methods can be categorized based on their ability to be model-agnostic. Some approaches are versatile and applicable to various learning problems and base models, while others are specifically designed for tasks like classification. See Section ~\ref{sec:relatedWork} and Table ~\ref{tab:few_shot_lit_review} in the Appendix for a detailed overview of this categorization.

For our few-shot regression setting in heterogeneous tasks, we need a model-agnostic approach that can handle regression and an amortization-based method that only adapts a few parameters to encourage parameter sharing between tasks and to prevent overfitting with fewer data. An example of such an approach is presented in \cite{abdollahzadeh2021revisit}. Similar to our approach TAM-RL, this modification shifts the meta-learner to an amortization-based structure (like ProtoNet \cite{snell2017prototypical}), which learns a metric space for comparing samples. Other methods ~\cite{oreshkin2018tadam,vinyals2016matching,abdollahzadeh2021revisit} also learn a similar metric space for comparing samples as the meta-learner. However, the use of similarity metric space restricts these methods to only classification tasks, and adopting them for regression-based tasks is nontrivial. A model with design choices closely related to TAM-RL is CNAP ~\cite{requeima2019fast}. CNAP employs and adapts a few sets of parameters in an amortized fashion. However, it uses a complex auto-regressive structure and expensive procedure for adaptation, making it infeasible for complex base models like recurrent neural networks. TAM-RL offers adaptability while maintaining computational efficiency and generalization capabilities by modulating only a few network parameters. Further, TAM-RL reduces adaptation costs due to its amortization-based approach. While this study is focused on regression, TAM-RL is a model-agnostic approach offering flexibility for different problem types.

%% file: 3_problemFormulation.tex
\section{Problem Formulation}
\label{sec:problemFormulation}

This study focuses on using ML models to learn driver-response behavior for various sparsely observed multi-modal (heterogeneous) entities. We have access to a collection of driver/response pairs of time series sequences for each entity $i$ in the train set. For each entity $i$ in the Training set, we can access a set of multivariate time series instances of corresponding inputs and output pairs, $D_i^{Train}$ = $[(\boldsymbol{X_i^1},y_i^1), (\boldsymbol{X_i^2}, y_i^2), \dots, (\boldsymbol{X_i^{T_{Train}}},y_i^{T_{Train}})]$, where the daily drivers $\boldsymbol{X_i}$ are a multivariate time series, where $\boldsymbol{X_i} = [\boldsymbol{x^1_i}, \boldsymbol{x^2_i}, \dots, \boldsymbol{x^T_i}]$ where, $\boldsymbol{x^t_i} \in \mathbb{R}^{D_x}$ represents the input vector at time $t \in T$ with $D_x$ dimensions, and $\boldsymbol{Y_i} = [y^1_i, y^2_i, \dots, y^T_i]$ represents the corresponding output. Additionally, for the set of entities in the test set, a few-shot data of inputs and outputs, $D_j^{Few}$ = $[(\boldsymbol{X_j^1},y_j^1), (\boldsymbol{X_j^2},y_j^2), \dots, (\boldsymbol{X_j^{T_{Few}}},y_j^{T_{Few}})]$ is provided for each entity $j$. The goal is to learn a regression function $\mathcal{F}: X \rightarrow Y$ that maps the input drivers to the output response for each entity in the test set.

Following the meta-learning literature, we split the data for each entity into two sets: a support set ($D_i^{support}$) and a query set ($D_i^{query}$). For the entities used in training, both sets are taken from the training data ($D_i^{Train}$). Whereas, for the entities in the test set, the few shot data ($D_j^{Few}$) is used as the support set. During training, meta-learning approaches use the support set ($D_i^{support}$) to understand the temporal correlations, multivariate relationships, and task characteristics associated with each entity ($i$). This information is then used for prediction in the query set ($D_i^{query}$). The objective is to train the model on $D_i^{Train}$ by minimizing the prediction error:
\vspace{-0.1in}
{\small
\begin{equation}
    \begin{split}
    \arg \min_{\theta} & \sum_{i} ||y_i^t - \mathcal{F}(\boldsymbol{x^t_i};D^{support})||^2\\
    where & \quad (\boldsymbol{x^t_i}, y_i^t) \in D_i^{query}
    \end{split}
\end{equation}
}

\subsection{Motivating Problems} 
\label{sec:motivating}
We apply our proposed framework to two societally important environmental problems: predicting Gross Primary Product (GPP) for flux towers with limited observations and forecasting streamflow in basins with sparse data. However, high-quality data for both carbon flux and streamflow is often limited for most parts of the world. This scarcity impedes our understanding of these critical environmental processes, particularly in regions with sparse data coverage.

%% file: 4_method.tex
\vspace{-0.1in}
\section{Method}
\label{sec:method}
Given sufficient training data, individual ML models can be trained for each entity. However, this is not feasible for entities that lack sufficient training data. The driver-response relationship for an entity is often governed by its inherent characteristics ($z_i$). Thus, the forward model is represented as $\mathcal{F}_{\theta}(x_i^t,z_i)$, where $\theta$ denotes the function class shared by the target systems and $z_i$ denotes entity-specific inherent characteristics. Our proposed Task Aware Modulation using Representation Learning (TAM-RL) framework builds a predictive model in a few-shot setting, for a set of entities for which heterogeneity is induced by unknown entity-specific characteristics. 
TAM-RL has two major components: \textit{modulation network} and \textit{base network}. The \textit{modulation network} consists of a BiLSTM-based \textit{task encoder} $\mathcal{E}$ that learns task-specific features representing the entity's characteristics, and a \textit{modulation parameter generator} $\mathcal{G}$ which predicts the modulation parameters using them. The \textit{base network} $\mathcal{F}_{\theta}$ is modulated by these parameters to generate modulated meta-parameters for the \textit{base network}. The modulated \textit{base network} uses the driver sequences to generate response sequences. TAM-RL further uses an adaptation method that adjusts the modulated base network to the specific task during inference. The following sections provide details on the neural network architectures used for TAM-RL and its training process.

\vspace{-0.1in}
\subsection{Modulation Network:}
\begin{figure}
    \centering
    \includegraphics[width=\columnwidth]{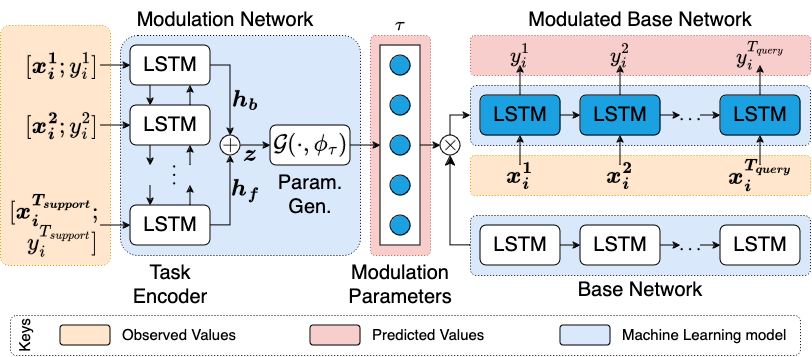}
    \caption{\footnotesize \textit{TAM-RL Architecture. The modulation network generates task-specific modulation parameters $\tau$, which are used to adapt the base network. This diagram illustrates the architecture when the base network is an LSTM.}}
    \label{fig:architecture}
    \vspace{-0.2in}
\end{figure}

We use a Bidirectional Long Short-Term Memory (LSTM)~\cite{graves2005framewise} network as the \textit{task encoder} to capture the temporal information and relationship between the driver and response in sequences. This step encodes the interaction between driver and response data for an entity to generate entity-specific embeddings. We add the last hidden states from both the forward ($\boldsymbol{h_f}$) and backward LSTMs ($\boldsymbol{h_b}$) in the \textit{task encoder} $\mathcal{E}$ to infer the final entity embeddings ($\boldsymbol{z}$), as depicted in Fig~\ref{fig:architecture}. These task embeddings capture the temporal information and interaction between the driver and the response for a given task.

\begin{equation}
    \label{eq:task encoder}
    \boldsymbol{z} = \boldsymbol{\mathcal{E}}([x^t;y^t]_{1:T};\phi_{z})
\end{equation}
where, $\boldsymbol{\mathcal{E}}$  is BiLSTM with parameters $\phi_z$.

The inferred entity embeddings are used to generate task-specific modulation parameters $\tau$ through an MLP-based \textit{modulation parameter generator} $\mathcal{G}$.
\begin{equation}
    \label{eq:modulation parameter generator}
    \begin{split}
        \boldsymbol{\tau} &= \boldsymbol{\mathcal{G}}(\boldsymbol{z};\phi_{\tau})
    \end{split}
\end{equation}
where, $\boldsymbol{\mathcal{G}}$  is an MLP with parameters $\phi_{\tau}$.

\vspace{-0.1in}
\subsection{Base Network:}
This base network is analogous to forward models in the context of environmental modeling. TAM-RL, being model agnostic, is compatible with any base model that can be trained with gradient descent and thus is a customized architectural choice unique to each domain and dataset. We use a sequence-to-sequence LSTM-based base network $\mathcal{F}$ that allows for the conditional generation of sequential responses given the sequential driver data, as shown in the base network block of Fig~\ref{fig:architecture}. During inference with TAM-RL, the modulation network is used to infer task-specific meta initialization of $\mathcal{F}$.

\vspace{-0.1in}
\subsection{Modulation:}
The modulation step plays a crucial role in TAM-RL being effective and parameter-efficient in a few-shot learning setting. We modulate our LSTM-based base network at two key points: (a) the input layer and (b) the final hidden state. Compared to the unconstrained modulation of MAML-based approaches, this approach requires only a small number of task-specific parameters, making it computationally efficient and less prone to overfitting. We perform the modulation via FiLM layer~\cite{perez2018film} using task-specific modulation parameters $\tau$ obtained from the encoder as shown below:
\vspace{-0.05in}
\begin{equation}
    \label{eq:modulation}
    \begin{split}
        \gamma_i^1, \beta_i^1 = f^{\text{input}}_{\text{task param}}(\tau_i) \qquad \boldsymbol{x^t_i}_{\text{mod}} &= \gamma_i^1 \odot f_d(\boldsymbol{x^t_i}) + \beta_i^1 \quad\\
        \gamma_i^2, \beta_i^2 = f^{\text{hidden}}_{\text{task param}}(\tau_i)\qquad \boldsymbol{h^t_i}_{\text{mod}} &= \gamma_i^2 \odot \mathbf{h}_i + \beta_i^2 \quad  
    \end{split}
\end{equation}

where $f_d$, $f^{\text{input}}_{\text{task param}}$, and $f^{\text{hidden}}_{\text{task param}}$ are linear transformations, $\odot$ denotes element-wise multiplication and $\mathbf{h}$ denotes the LSTM output. The modulated base network is then optimized for the target task through gradient-based optimization, which learns all the modulation and base network parameters, similar to that of MMAML~\cite{vuorio2019multimodal}.

\subsection{Training} of TAM-RL occurs in two stages: pre-training \textit{base network} and joint training of \textit{modulation parameter generator} and modulated \textit{base network}.

\noindent\textbf{Pre-training:} In the first stage, we pre-train the initial parameters $\theta$ of the base network (forward model) using all entities in the dataset without incorporating any site-specific information. This pre-training step acknowledges that while all entities belong to the same broad task category (e.g., streamflow prediction or GPP estimation), they exhibit significant heterogeneity in their characteristics and behaviors. We employ a full-way regression procedure, where the model is trained to map inputs directly to outputs across the entire dataset without incorporating any site-specific information. This approach allows the base network to capture a wide range of patterns and relationships present across the heterogeneous entities while capturing the general structure and patterns common to all entities within the task distribution. By initializing the model with these pre-trained weights, we establish a versatile foundation that can accommodate the diversity within the task distribution. This pre-training strategy provides a starting point that is exposed to the full spectrum of entity behaviors, preparing the model for subsequent task-specific adaptation and modulation in the meta-learning framework.  This approach mirrors standard pre-training techniques and allows the base network to capture broad patterns and dependencies present across different entities. By initializing the model with these pre-trained weights, we provide a strong foundation that encodes shared knowledge across the distribution for subsequent task-specific adaptation in the meta-learning framework.

\noindent\textbf{Joint training:} To overcome the challenge of feeding a long time series into a model, we divide the training data ($D^{Train}$) into $\mathcal{T}$ sliding windows ($W^{Tr_{\mathcal{T}}}$) during the training phase. These windows are further split into support windows ($W^{support_{\mathcal{T}}}$) and query windows ($W^{query_{\mathcal{T}}}$) for training the model. For each task in the support set, we select a support window ($W^{support_{\mathcal{T}}}$) from the Train($W^{Tr_{\mathcal{T}}}$) sliding windows and extract the corresponding response and driver data for the given entity. The query set is constructed similarly, with the remaining data that was not used to create the support set. This ensures that the model is trained on a diverse set of support tasks and can generalize to unseen tasks during inference. The entire framework is trained end to end using a forward loss jointly across all training tasks, which can be any supervised loss, depending on the problem. In our experiments, we use mean squared error (MSE) as the supervised loss (Eq ~\ref{eq:mse_loss}). 
\vspace{-0.05in}
\begin{equation}
    \mathcal{L} = \frac{1}{N} \sum_{i=1}^N \frac{1}{T} \sum_{t=1}^T (y_i^t-\hat{y}_i^t)^2
    \label{eq:mse_loss}
\end{equation}

The final objective of TAM-RL during training can be formally expressed as:
{\small
\begin{equation}
    \begin{split}
    \arg \min_{\theta , \phi_{h}, \phi_{\tau}} \sum_{i} ||y_i^t - \mathcal{F}(\boldsymbol{x^t_i};\theta^{'}_{i})||^2 \quad where \quad \theta^{'}_{i} = \theta_{i} \bigotimes \tau, \\ 
    \tau = \mathcal{G}(z; \phi_t), \\
    \boldsymbol{z} = \mathcal{E}(W_i^{support_\tau},\phi_h ),\\
    s.t. (\boldsymbol{x^t_i}, y_i^t) \in W_i^{query_\tau}
    \end{split}
\end{equation}
}%
\vspace{-0.1in}

The pseudocode for the training process can be found in the Appendix at Algorithm~\ref{alg:training_pseudo}.

\subsection{Inference:} In the inference phase, we use a small amount of data ($D_j^{few}$) for a new entity $j$, which we use to create $\mathcal{T}$ support sliding windows ($W_j^{support_{\mathcal{T}}}$). These windows, along with the input data ($x_j^{query_{\mathcal{T}}}$), are used to make predictions. To do this, we first run a modulation network on a small amount of data to learn task-specific modulation parameters that modulate the base network. The modulated base network is then adapted using a few shots of data. Finally, we use the task-specific characteristics, the input data, and the adjusted model parameters to make predictions. The detailed procedure is outlined in Algorithm~\ref{alg:inference} in the Appendix. The modulated base network is further optimized for the target task through gradient-based optimization, similar to MMAML~\cite{vuorio2019multimodal}. We use an ADAM-based optimizer for the Adaptation step. It should be noted that the modulation network remains unaltered during the inference phase, preventing any negative impact from adaptation.

\subsection{Advantages of TAM-RL:} TAM-RL offers several key advantages, specifically in the context of environmental systems. First, the task-specific forward and inverse models in TAM-RL are trained in an amortized fashion (where the forward model is just a forward pass). This amortized training enables this framework to learn a high-quality \textit{modulation network} due to the absence of an adaptation step. Second, by replacing inner-loop adaptation with an amortization (forward pass) of the base network, TAM-RL achieves enhanced gradient propagation, leading to a better-quality modulation network. Third, in TAM-RL, the model is learned jointly across all tasks and not in a task-wise episodic fashion as in common meta-learning approaches, allowing it to learn the relationships between tasks~\cite{hospedales2021meta}. Consequently, this design choice makes TAMRL more stable across architectures and hyperparameters, faster during training, requires minimal hyperparameter tuning, and avoids the need for second derivatives.

%% file: 5_experiments.tex
\section{Experiments}
\label{sec:experiments}

\input{tables/dataset}


\subsection{Datasets:} 
\begin{figure}[t]
    \centering
    \includegraphics[width=\columnwidth]{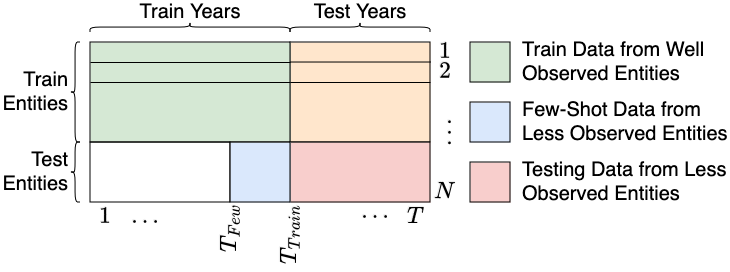}
    \caption{\footnotesize \textit{Few-shot prediction setting followed in this paper.}}
    \label{fig:experiment setting}
\end{figure}

We evaluate TAM-RL's effectiveness for a few-shot prediction setting (see Figure~\ref{fig:experiment setting}) using two publicly available real-world environmental datasets and a controlled synthetic dataset.

\noindent\textbf{FLUXNET}~\cite{pastorello2020fluxnet2015} consists of data from a global network of eddy-covariance stations that record carbon, water, and energy exchanges between the atmosphere and the biosphere, covering a wide range of climate and ecosystem types. In this study, the task is to estimate Gross Primary Productivity (GPP) on a daily timescale using meteorological and remote sensing inputs such as precipitation, air temperature, and vapor pressure. We selected 150 towers with over three years of data and split them into training and few-shot testing sites, as shown in the descriptive statistics in Table~\ref{tab:dataset}.

\noindent\textbf{CARAVAN-GB}~\cite{kratzert2022caravan} is a subset (focusing on UK) of the CARAVAN dataset~\cite{kratzert2022caravan} which is a global benchmark dataset comprising of basins from various regions across the world. In this study, the task is to estimate streamflow on a daily timescale using meteorological forcing data (e.g., precipitation, evaporation, temperature) and basin characteristics. We selected 255 UK basins with ten years of streamflow observations (1989-1999), splitting them into training and few-shot testing sites as shown in the descriptive statistics in Table~\ref{tab:dataset}.

\noindent\textbf{Synthetic Data}~\cite{vuorio2019multimodal} is a set of task distributions for regression with several collections of functions called modes. We follow the setup used by Vuorio et al.~\cite{vuorio2019multimodal} and include the following $five$ modes: \textit{sine}, \textit{linear}, \textit{quadratic}, \textit{$\ell_1$ norm} and \textit{tanh}. We create three sets of multi-modal task distribution (in decreasing order of task heterogeneity, $\text{SET}_1>\text{SET}_2>\text{SET}_3$), each set consisting of three modes: $\mathbf{SET_1}$ \{$sine$, $linear$, $quadratic$\}, $\mathbf{SET_2}$ \{$linear$, $tanh$, $\ell_1 norm$\}, and 
$\mathbf{SET_3}$ \{$quadratic$, $tanh$, $\ell_1 norm$\}

We select 375,000 tasks for each of the three modes, each with $five$ meta-train and $five$ meta-validation examples for training. We sample 12,500 new tasks for each mode to evaluate each set with $five$ few-shot samples.

We direct the reader to Appendix~\ref{sec:dataset_details} for additional details on the source of the datasets and the steps required to prepare them for the experiments shown in this paper.

\subsection{Baselines:} In the experiments, we evaluate the effectiveness of TAM-RL for few-shot prediction tasks by comparing it with several representative meta-learning methods. \noindent\textbf{ProtoNet}~\cite{snell2017prototypical} is a modification over the original method (designed for classification) to make it model-agnostic and suitable for regression tasks. Our modified version combines task-specific local models $F_{\theta_j}$ learned from the support set S with a global embedding function $\phi: \mathbb{R}^d \rightarrow \mathbb{R}^m$. For a query point $x_i$, the prediction is defined as: $y_i = \sum_{(x_j, y_j) \in S} w_{ij} \cdot F_{\theta_j}(x_i)$, where $w_{ij} = \frac{\exp{\langle\phi(x_j), \phi(x_i)\rangle}}{\sum_{(x_k, y_k) \in S} \exp{\langle\phi(x_k), \phi(x_i)\rangle}}$ similar to that in ~\cite{lu2021towards}. \textbf{Meta-Transfer Learning} (MTL)~\cite{ghosh2022meta} estimates the similarity amongst entities to transfer knowledge from well-observed entities to unmonitored entities. We use a similar architecture as TAM-RL, with the similarity measured using the embeddings from the sequential encoder. \textbf{MAML}~\cite{finn2017model} uses the same base network (Decoder) in TAM-RL. \textbf{MMAML}~\cite{vuorio2019multimodal} trains a global model, where we use a BiLSTM and MLP for the \textit{task encoder} and \textit{modulation parameter generator} in the \textit{modulation network} and LSTM as the \textit{base network}. \textbf{KGSSL}~\cite{ghosh2022robust} is the state-of-the-art purely deterministic inverse framework~\cite{ghosh2022robust} for few-shot settings to infer the entity attributes in the form of embeddings and further use them to predict the streamflow. Lastly, only for comparison, we also present results using \textbf{CTLSTM}~\cite{kratzert2019towards} and \textbf{CTLSTM$_{\text{finetune}}$}, both of which have access to the actual basin characteristics not used in our proposed method.

\vspace{-0.1in}
\subsection{Implementation Details:} To make a fair comparison, we use a similar configuration for the baselines and TAM-RL, including neural network layers, hidden units, regularization, and parameter initialization. For all the methods, we use Adam optimizer~\cite{kingma2014adam} for training and select optimal values of hyperparameters on each dataset based on their performance in the validation set. To mitigate the impact of randomness due to network initializations, we report the results of the ensemble prediction obtained by averaging the predictions of \textit{five} models with different weight initializations for all methods. More details on model implementation and hyperparameter selection are provided in Appendix~\ref{sec:experimental_setting}.

%% file: tables/dataset.tex
\begin{table}[t!]
    \caption{\footnotesize \textit{Descriptive statistics of real-world datasets. Additional dataset details are provided in the supplementary section.}}
    \label{tab:dataset}
    \vspace{-0.1in}
    \resizebox{\columnwidth}{!}{
        \begin{tabular}{|c|c|c|c|c|}
            \hline
            \multirow{2}{*}{Dataset} & \multicolumn{2}{c|}{Train Setting}   & \multicolumn{2}{c|}{Test Few-shot Setting} \\
            \cline{2-5}
                                     & N                             & K                              & N             & K \\
            \hline
            FLUXNET                  & 145 sites                           & 3 years                             & 15 sites      & 1, 3, 12, 24 months\\
            CARAVAN-GB               & 204 basins                          & 10 years                           & 51 basins     & 1, 2, 5 years \\
            Synthetic                & 112500 functions                    & 10         & 37500 functions     & 5\\
            \hline
        \end{tabular}
    }
    \vspace{-0.25in}
\end{table}

%% file: 6_results.tex
\section{Results}
\label{sec:results}

\input{tables/few-shot_real-world_evaluation}

\subsection{Few-shot Evaluation on Real World Datasets:} This section presents the experimental results on two real-world datasets. Table~\ref{tab:few-shot_real-world_evaluation} shows the mean ensemble Root Mean Square Error (RMSE) values for model predictions. We evaluated several methods across different few-shot learning scenarios (see Figure \ref{fig:experiment setting}), ranging from 1 to 24 months for FLUXNET and 1 to 5 years for CARAVAN-GB. For the GPP prediction in FLUXNET2015, the LSTM base network processes input sequences of 30 days with an output stride of 15 days. In CARAVAN-GB, we create sliding windows of 365 days, strided by half the sequence length, i.e., 183 days.

From Table~\ref{tab:few-shot_real-world_evaluation}, our proposed TAM-RL method consistently outperformed the representative meta-learning methods, i.e., MAML, ProtoNet, and MTL, across all few-shot scenarios in both the datasets. Moreover, it outperforms advanced meta-learning algorithms like MMAML and KGSSL which is especially designed for environmental tasks. The performance improvement for the FLUXNET dataset was particularly notable in the 12-month and 24-month scenarios, where TAM-RL significantly outperforms the next best method, MMAML, by 13.56\% and 10.33\%, respectively. For the CARAVAN-GB dataset, the improvement was most pronounced in the 5-year scenario, where TAM-RL outperforms KGSSL and MMAML by 6.67\% and 10.06\%, respectively. Note that, although CTLSTM and its fine-tuned version, CTLSTM$_\text{finetune}$ have additional information in the form of the entity's inherent characteristics, both are outperformed by TAM-RL in most cases.

\subsection{Ablation Studies}
\label{sec:ablation}

We conduct ablation studies to evaluate the effectiveness of pre-training, modulation, and adaptation mechanisms on the performance of TAM-RL. Table \ref{tab:few-shot_real-world_ablation} presents the Mean Ensemble Root Mean Square Error (RMSE) for GPP prediction on the FLUXNET dataset and streamflow modeling on the CARAVAN-GB Benchmark dataset, comparing different configurations of TAM-RL.

\noindent\textbf{Effect of Pre-training:} To assess the impact of pre-training the base network, we compared TAM-RL with and without pre-training (denoted by the superscript ``no pre-train"). For the FLUXNET dataset, pre-training showed a consistent improvement across all few-shot scenarios. Specifically, the RMSE decreases from 1.91 to 1.78 with 12 months of data and from 1.83 to 1.75 with 24 months of data when pre-training is included. Similarly, for the CARAVAN-GB dataset, the RMSE improves from 1.19 to 1.16 when 5 years of data is used. Thus, the results show that pre-training provides a significant boost in performance across both datasets.

\input{tables/few-shot_real-world_ablation}

\noindent\textbf{Effect of Modulation and Adaptation:} To evaluate the impact of adaptation, we compared TAM-RL with and without fine-tuning (TAM-RL$_{\text{no finetune}}$). In the FLUXNET dataset, TAM-RL$_{\text{no finetune}}$  achieves an RMSE of 2.28 with three months of data, whereas the fully adaptive TAM-RL improves this to 2.23. In the CARAVAN-GB dataset, adaptation leads to an improvement from 1.26 to 1.23 with two years of data and from 1.24 to 1.16 with five years of data. These results suggest that fine-tuning the model on the specific task data further refines the predictions, especially when more task-specific data is available. Further, we observe that TAM-RL$_{\text{no finetune}}^{\text{no pretrain}}$  performs better or on par with the next best baselines in these two datasets. While part of the improvement in TAM-RL can be attributed to a better base network, the majority of TAM-RL's performance gains stem from the modulation of task-specific parameters (TAM-RL$_{\text{no finetune}}$), leading to a comparable or better performance than other baselines on real-world datasets without fine-tuning. This is evident from GPP prediction, where TAM-RL with less than or equal to 3 months of few-shot data outperforms all baselines. This can be attributed to TAM-RL's modulation network (inverse model), which is trained to map directly from the few-shot data to the modulated parameters. This incentive motivates them to capture more efficient task-specific modulations, resulting in a more accurate inverse model. Moreover, both TAM-RL and TAM-RL$_{\text{no finetune}}$ performance improves as the quantity of few-shot data increases, demonstrating the strength of its modulation network. Additionally, for both GPP prediction (FLUXNET) and streamflow prediction (CARAVAN-GB), TAM-RL$_{\text{no finetune}}$ demonstrates a trend of performance improvement as the number of few-shot years of data increases. This illustrates TAM-RL's capacity to effectively leverage additional data points, resulting in improved modulation parameters.


\input{tables/few-shot_synthetic_evaluation}
\vspace{-0.1in}
\subsection{Few-shot Evaluation on Synthetic Dataset} 

The design choice of TAM-RL, which leverages meta-initializations directly through an amortization approach, proves highly effective in generating superior meta-initializations for diverse tasks through learned distinct representations. To rigorously evaluate, we designed a controlled synthetic data experiment by systematically varying the degree of task heterogeneity. We train MAML, MMAML, and TAM-RL for each of the three sets of multimodal distributions, SET$_1$, SET$_2$, and SET$_3$ mentioned in Section~\ref{sec:experiments}. From $SET_1$ to $SET_3$, the class of functions becomes increasingly homogeneous. Specifically, Quadratic functions can resemble sinusoidal or linear ones, but sinusoidal and linear functions differ. $SET_3$ is homogeneous as quadratic, $\ell_1$, and hyperbolic tangent functions are similar, especially with Gaussian noise. Following~\cite{vuorio2019multimodal}, we use an MLP-based network for three best-performing baseline models (MAML, MMAML, TAM-RL), with MMAML and TAM-RL using Bi-LSTM as task encoder (refer to ~\ref{sec:experimental_setting} for details of the architecture). We use ADAM as the meta-optimizer with the same hyperparameters from ~\cite{vuorio2019multimodal}, using five gradient steps and five meta-train examples during the evaluation. We report the MSE of five meta-validation examples for new tasks.

We report the mean MSE for each architecture on all three sets in Table~\ref{tab:few-shot_synthetic_evaluation}. For SET$_1$, the performance of TAM-RL is 21.6\% better than MMAML. As the set becomes more homogeneous, like in SET$_2$, the performance gap between TAM-RL and MMAML narrows and, in this scenario, TAM-RL only holds a slight edge of 3.5\% over MMAML, and they operate at a similar performance level. When the dataset reaches near complete homogeneity, as in SET$_3$, TAM-RL performs similarly to MMAML, with TAM-RL being worse by 5.9\%. Our results show that TAM-RL outperforms MMAML as the set becomes more heterogeneous. This is likely because TAM-RL relies heavily on finding distinct task-specific parameters to produce an effective modulated base network, which becomes more challenging as tasks become increasingly less distinguishable. Additionally, from Table~\ref{tab:few-shot_synthetic_evaluation}, we can observe that MAML has the highest error in all settings and that incorporating task identity through task-specific parameters significantly matters in multi-modal task distributions.

\begin{figure}
    \vspace{-0.2in}
    \centering
    \includegraphics[width=\linewidth]{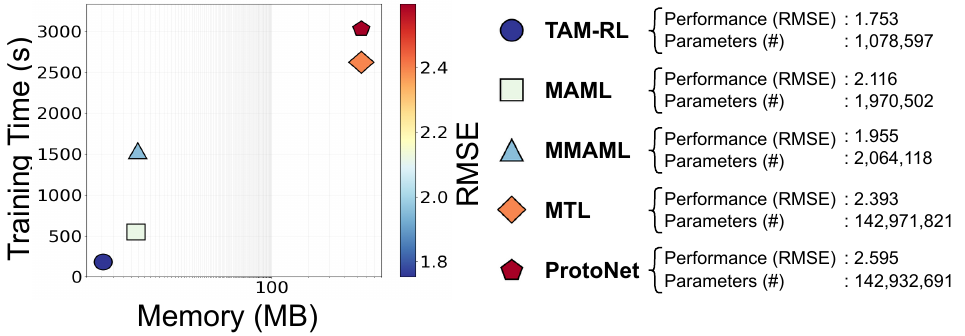}
    \caption{\footnotesize \textit{Performance comparison of various models regarding memory usage, training time, and RMSE on the Fluxnet dataset.}}
    \vspace{-0.3in}
    \label{fig:model_comparison}
\end{figure}
\subsection{Computational Space and Time Analysis}
Experimental results on the Fluxnet dataset for 24 months few shot setting on an Nvidia A-100 GPU demonstrate TAM-RL's significant advantages in computational efficiency, memory utilization, and performance compared to other meta-learning approaches such as MAML, MMAML, Meta Transfer Learning, and ProtoNet \ref{fig:model_comparison}. TAM-RL achieves a 3-fold to 14-fold reduction in training time, attributed to its amortized training approach requiring only one forward and backward pass per batch, in contrast to the multiple forward passes needed by gradient-based methods. Furthermore, TAM-RL maintains the lowest memory footprint among all compared methods, using 2 to 5 times less memory than its closest competitors and over 100 times less than the most memory-intensive approaches. Notably, despite its computational and memory efficiency, TAM-RL demonstrates superior performance with the lowest RMSE among all evaluated methods. Furthermore, TAM-RL offers advantages in optimization complexity with fewer crucial hyperparameters, potentially reducing the resources required for tuning. 






%% file: tables/few-shot_real-world_evaluation.tex
\begin{table}[]
    \caption{\footnotesize \textit{Root Mean Square Error (RMSE) for GPP prediction (FLUXNET) and streamflow prediction (CARAVAN-GB). The amount of data used as few-shots are denoted as column names.}}
    \label{tab:few-shot_real-world_evaluation}
    \vspace{-0.1in}
    \resizebox{\columnwidth}{!}{
        \begin{tabular}{|c|cccc|ccc|}
            \hline
            \multirow{2}{*}{Methods} & \multicolumn{4}{c|}{FLUXNET (in Months)}                                                   & \multicolumn{3}{c|}{CARAVAN-GB (in years)}                          \\ \cline{2-8} 
                                     & 1                    & 3                    & 12                   & 24                    & 1                    & 2                    & 5                     \\ \hline
            ProtoNet                 & 2.842                & 2.747                & 2.595                & 2.598                 & 1.432               & 1.427                & 1.429                 \\
            MTL                     & 3.066                & 2.592                & 2.393                & 2.401                 & 1.469               & 1.418               & 1.418                 \\
            MAML                     & 2.709                & 2.543                & 2.129                & 2.116                 & 1.401                & 1.386                & 1.384                 \\
            MMAML                    & 2.773                & 2.649                & 2.065                & 1.955                 & 1.348                & 1.339                & 1.292                 \\
            KGSSL                    & 2.366                & 2.349                & 2.237                & 2.205                 & 1.473                & 1.375                & 1.245                 \\
            TAM-RL                   & \textbf{2.316}       & \textbf{2.228}       & \textbf{1.785}       & \textbf{1.753}        & \textbf{1.245}       & \textbf{1.229}       & \textbf{1.162}        \\
            \hdashline
            CTLSM                    & 2.232                & 2.232                & 2.232                & 2.232                 & 1.364                & 1.364                & 1.364                 \\
            CTLSTM$_{finetune}$         & 2.324                & 2.267                & 1.756                & 1.621                 & 1.247                & 1.201                & 1.189                 \\ \hline
        \end{tabular}
    }
    \vspace{-0.2in}
\end{table}

%% file: tables/few-shot_real-world_ablation.tex
\begin{table}[]
    \caption{\footnotesize \textit{Root Mean Square Error (RMSE) for GPP prediction (FLUXNET) and streamflow prediction (CARAVAN-GB). The amount of data used as few-shots is denoted as column names.}}
    \label{tab:few-shot_real-world_ablation}
    \resizebox{\columnwidth}{!}{%
        \begin{tabular}{|l|cccc|ccc|}
        \hline
        \multirow{2}{*}{Methods} & \multicolumn{4}{c|}{FLUXNET (in Months)}                      & \multicolumn{3}{c|}{CARAVAN-GB (in years)}    \\ \cline{2-8} 
                                       & 1    & 3    & 12   & 24   & 1    & 2    & 5    \\ \hline
        TAM-RL$_{\text{no finetune}}^{\text{no pretrain}}$ & 2.32 & 2.31 & 2.21 & 2.18 & 1.45 & 1.29 & 1.25 \\
        TAM-RL$_{\text{no finetune}}$             & 2.32 & 2.28 & 2.14 & 2.11 & 1.39 & 1.26 & 1.24 \\
        TAM-RL$^{\text{no pretrain}}$             & 2.33 & 2.31 & 1.91 & 1.83 & 1.29 & 1.23 & 1.19 \\
        TAM-RL                   & \textbf{2.32} & \textbf{2.23} & \textbf{1.78} & \textbf{1.75} & \textbf{1.24} & \textbf{1.23} & \textbf{1.16} \\ \hline
        \end{tabular}%
    }
    \vspace{-0.2in}
\end{table}

%% file: tables/few-shot_synthetic_evaluation.tex
\begin{table}[t]
    \caption{\footnotesize \textit{Mean square error (MSE) on the multimodal 5-shot regression with different combinations of three modes for different architectures across different sets. Gaussian noise is applied to each function with $\mu=0$ and $\sigma=0.3$. Values in brackets represent the standard deviation across tasks in the set. See Table~\ref{tab:expanded_sythetic_table} for expanded version of this table}}
    \label{tab:few-shot_synthetic_evaluation}
    \centering
    \resizebox{\columnwidth}{!}{
        \begin{tabular}{|l|l|l|l|}
            \hline
            \textbf{Architecture} & \textbf{SET$_1$} & \textbf{SET$_2$} & \textbf{SET$_3$} \\ \hline
            MAML     & 3.684 (0.176)          & 3.446 (0.115)         & 1.454 (0.040)         \\ 
            MMAML    & 0.601 (0.074)          & 1.085 (0.121)         & \textbf{0.757 (0.05)} \\ 
            TAM-RL (ours) & \textbf{0.494 (0.040)} & \textbf{1.049 (0.69)} & 0.805 (0.023) \\ \hline
        \end{tabular}
    }    
    \vspace{-0.2in}
\end{table}

%% file: 7_conclusion.tex
\section{Conclusion}
\label{sec:conclusion}

This paper presented TAM-RL, a novel task-aware modulation framework using representation learning, for use in a few-shot setting. This approach is particularly suited for environmental applications with an underlying forward model common to all entities where heterogeneity arises due to entity/task characteristics. We performed extensive experiments on two real-world environmental datasets, a GPP predictions task for flux towers and a hydrological benchmark dataset CARAVAN GB, showing that this framework TAM-RL outperforms baseline models for less-observed entities.  We further showed that TAM-RL outperforms MMAML when multimodal task distribution is more heterogeneous.  Our proposed method is general and can add value to domains beyond environmental applications. Our framework cannot handle missing driver or response data observations. However, this methodology can be further extended to handle missing observations. Another potential direction is to improve the modulation network(Inverse model) by incorporating Knowledge-guided Self-Supervised Learning\cite{ghosh2022robust} into TAM-RL's task encoder,  which has the potential to make the task parameters semantically meaningful.

\section{Acknowledgement}
This work was supported by the NSF LEAP Science and Technology Center (award 2019625) and the NSF grant (2313174). Computational resources were provided by the Minnesota Supercomputing Institute.

%% file: Appendix.tex
\begin{center}
\section*{Appendix}
\end{center}

\section{Related Work}
\label{sec:relatedWork}

\begin{figure}[h]
    \centering
    \includegraphics[width=0.9\linewidth]{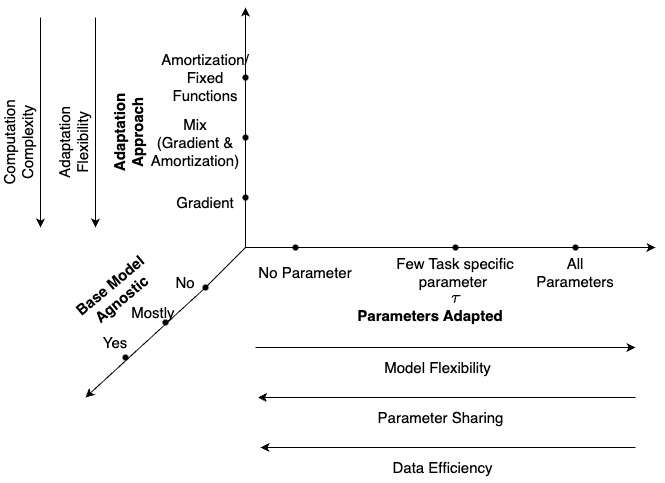}
    \caption{\footnotesize \textit{Categorization of meta-learning approaches. The X-axis represents Parameters Adapted, the Y-axis represents the Adaptation Approach, and the Z-axis represents whether the Base Model is Agnostic. All Arrows go from low to high.}}
    \label{fig:meta_learning_grouping}
\end{figure}
Most recent successful approaches for few-shot learning use meta-learning to enable models to learn quickly from limited examples and adapt to new tasks efficiently. There is extensive literature on meta-learning \cite{hospedales2021meta,huisman2021survey}. Extending from \cite{requeima2019fast}'s work, we categorize existing methods based on the following three criteria as illustrated in Figure~\ref{fig:meta_learning_grouping}
\begin{enumerate}
    \item Parameters Adapted: How many of the model parameters are adapted for each task?
    \item Adaptation Approach: How does the meta-model approach compute task-specific parameters from a few examples?
    \item Model Agnostic: How agnostic is the adaptation approach to different base models and learning tasks?
\end{enumerate}

\noindent \textbf{Parameters Adapted:}  
In few-shot learning across multiple tasks, adapting at least a few task-specific parameters of the model ensures adaptability to task-specific features, enhancing the model's capacity for effective generalization. Lots of meta-learning approaches do just this, such as \cite{snell2017prototypical, vinyals2016matching, gordon2018meta, triantafillou2019meta, oreshkin2018tadam,abdollahzadeh2021revisit,wang2020structured,requeima2019fast,jiang2018learning,zintgraf2019fast}. We can also adapt all parameters to increase our flexibility further like in the case of \cite{finn2017model,vuorio2019multimodal,arango2021multimodal,antoniou2018train,li2017meta,loo2019few,patacchiola2020bayesian,baik2021meta,sung2018learning,rusu2018meta}. Increasing the number of task parameters enhances flexibility and expressibility for task parameters but increases the risk of overfitting with limited data, less data efficiency, and parameter sharing.  In contrast, CNPs \cite{garnelo2018conditional} and KGSSl\cite{ghosh2022robust} represent an extreme case within this spectrum by not adapting any parameters and sharing all parameters across tasks, making them highly efficient. Further, CNPs define task-specific parameters $\tau$ as a fixed-dimensional vector used as an input to the model rather than as specific parameters of the model itself. However, this characteristic makes CNPs unsuitable for modeling heterogeneous distributions as adapting at least very few task-specific parameters of the model ensures adaptability to task-specific features, enhancing the model's capacity for effective generalization in heterogeneous settings.  There's a trade-off: more adaptable models (with many adjusted parameters) are flexible but computationally expensive and prone to overfitting. To balance this, TAM-RL only modulates a small portion of the network parameters, similar to \cite{oreshkin2018tadam,abdollahzadeh2021revisit,requeima2019fast}.

\noindent \textbf{Adaptation Approach:} Progressing from amortized-based approaches \cite{oreshkin2018tadam,snell2017prototypical,vinyals2016matching,abdollahzadeh2021revisit,gordon2018meta,requeima2019fast,garnelo2018conditional,loo2019few,patacchiola2020bayesian,sung2018learning,kim2021multi,ghosh2022robust} to gradient-based approaches\cite{finn2017model,antoniou2018train,li2017meta,wang2020structured,jiang2018learning,baik2021meta,zintgraf2019fast} brings more flexibility in the adaptation approach but comes with higher computational expenses and a propensity for overfitting, leading to increased complexity and instability. This, in turn, necessitates careful tuning for stable hyperparameters. While amortization substantially reduces the cost of adaptation, it may encounter challenges related to the amortization gap, resulting in underfitting, especially in large data regimes, as these methods involve training neural networks to directly produce task-specific parameter values. But as our focus is on a few-shot setting, TAM-RL uses an amortization-based approach. Recent work has also proposed a hybrid approach, incorporating both amortization and gradient-based methods, aiming to leverage the advantages of the amortization approach to gradient-based optimization to reduce their complexity and instability \cite{vuorio2019multimodal,arango2021multimodal,triantafillou2019meta,rusu2018meta}.

\noindent \textbf{Base Model Agnostic:} Approaches range from Model agnostic \cite{finn2017model,antoniou2018train,li2017meta,vuorio2019multimodal,wang2020structured,arango2021multimodal,gordon2018meta,garnelo2018conditional,jiang2018learning,loo2019few,patacchiola2020bayesian,baik2021meta,ghosh2022robust,kim2021multi,zintgraf2019fast,ghosh2022meta} which are applicable to various learning problems/ base models  (e.g., classification, regression, reinforcement learning) to approaches that are specifically designed for classification\cite{oreshkin2018tadam,snell2017prototypical,vinyals2016matching,abdollahzadeh2021revisit,sung2018learning,triantafillou2019meta,rusu2018meta,requeima2019fast}. As the focus of this paper is on regression, TAM-RL uses a model-agnostic approach that can handle regression.

\noindent \textbf{ Approaches most relevant for our setting }
For our few-shot regression setting in heterogeneous tasks, we need a model-agnostic approach that can handle regression and an amortization-based method that only adapts a few parameter parameters to prevent overfitting with less data. One of the most prominent meta-learning approaches in Model-Agnostic Meta-Learning (MAML) \cite{finn2017model} and its gradient-based adaptation variants (e.g., \cite{antoniou2018train,li2017meta,baik2021meta}) that use gradient-based adaptations to modify all parameters of the model from one initialization. These models use the most flexible approach for both adaptation and the parameters adapted; hence, they are more complex and often prone to overfitting. Further, this way of adaptation fails to generalize to multimodal distribution. To address the challenges posed by multimodal task distributions, extensions of MAML, such as those proposed in \cite{vuorio2019multimodal, wang2020structured, arango2021multimodal}, introduce multiple initializations. While these multimodal gradient-based methods successfully extend MAML, they still do not address the core challenges of a model with gradient-based optimization, such as the propensity to overfit due to flexibility. To address these challenges, an approach that shifts from gradient-based adaptation to amortization-based approaches has been proposed. An example of such an approach is presented in \cite{abdollahzadeh2021revisit}, which addresses gradient flow, computation, and stability issues associated with multimodal gradient-based methods. Similar to our approach TAM-RL, this modification shifts the meta-learner to an amortization-based structure. \cite{abdollahzadeh2021revisit} achieves this by transitioning from a MAML-trained base network to a simpler meta-learner, ProtoNet \cite{snell2017prototypical}, which learns a metric space for comparing samples. There also exist other methods ~\cite{oreshkin2018tadam,vinyals2016matching,abdollahzadeh2021revisit} that learn a similar metric space for comparing samples as the meta-learner. However, the use of similarity metric space restricts these methods to only classification tasks, and adopting them for regression-based tasks is nontrivial. A model with design choices closely related to TAM-RL is CNAP ~\cite{requeima2019fast}. CNAP employs and adapts a few sets of parameters in an amortized fashion. However, it uses a far more complex auto-regressive structure and expensive procedure for adaptation (i.e., an auto-regressive component that provides information to deeper layers in the feature extractor concerning how shallower layers have
adapted to the task), which makes it infeasible to use for complex base models (feature extractor) like recurrent neural networks.

See Table ~\ref{tab:few_shot_lit_review} in the Appendix for a summary of various few-shot learning methods and their categorization based on the Parameters Adapted, Adaptation Approach, and Base Model Agnostic properties.

\begin{table}
\centering
\resizebox{\columnwidth}{!}{%
\begin{tabular}{|l|l|l|l|}
\hline
\textbf{Paper name/number}                               & \textbf{Parameters Adapted} & \textbf{Adaptation Approach} & \textbf{Model Agnostic} \\ \hline
TAMRL (Our approach)                             & Few & Amortized & Yes \\ \hline
\cite{oreshkin2018tadam}        & Few & Amortized & No  \\ \hline
\cite{snell2017prototypical}    & Few & Amortized & No  \\ \hline
\cite{vinyals2016matching}      & Few & Amortized & No  \\ \hline
\cite{finn2017model}            & All & Gradient  & Yes \\ \hline
\cite{antoniou2018train}        & All & Gradient  & Yes \\ \hline
\cite{li2017meta}               & All & Gradient  & Yes \\ \hline
\cite{abdollahzadeh2021revisit} & Few & Amortized & No  \\ \hline
\cite{vuorio2019multimodal}     & All & Mix       & Yes \\ \hline
\cite{wang2020structured}       & Few & Gradient  & Yes \\ \hline
\cite{arango2021multimodal} & All & Mix       & Yes \\ \hline
\cite{gordon2018meta}           & Few & Amortized & Yes \\ \hline
\cite{requeima2019fast}         & Few & Amortized & Mostly  \\ \hline
\cite{garnelo2018conditional}   & No  & Amortized & Yes \\ \hline
\cite{jiang2018learning}        & Few & Gradient  & Yes \\ \hline
\cite{loo2019few} & All                                                 & Amortized                    & Yes                     \\ \hline
\cite{patacchiola2020bayesian}  & All & Amortized & Yes \\ \hline
\cite{baik2021meta}             & All & Gradient  & Yes \\ \hline
\cite{sung2018learning}         & All & Amortized & No  \\ \hline
\cite{triantafillou2019meta}    & Few & Mix       & No  \\ \hline
\cite{rusu2018meta}             & All & Mix       & No  \\ \hline
\cite{ghosh2022robust}          & No  & Amortized & Yes \\ \hline
\cite{kim2021multi}             & All & Amortized & Yes \\ \hline
\cite{zintgraf2019fast}         & Few & Gradient  & Yes \\ \hline
\cite{ghosh2022meta}
& All & Amortized & Yes \\
\hline
\end{tabular}%
}
\caption{\small Summarized overview of various few-shot learning methods and their categorization based on the Parameters Adapted, Adaptation Approach, and Model Agnostic properties}
\label{tab:few_shot_lit_review}
\end{table}

\section{Dataset Details}
\label{sec:dataset_details}
\subsection{ Fluxtower Dataset} 
The FLUXNET2015~\cite{pastorello2020fluxnet2015} dataset is a global network of eddy-covariance stations that record carbon, water, and energy exchanges between the atmosphere and the biosphere. It offers valuable ecosystem-scale observations across various climate and ecosystem types. In this dataset, our goal is to estimate Gross Primary Productivity (GPP) on a daily timescale for FLUXNET2015 sites. We achieve this by using a combination of meteorological and remote sensing inputs, including precipitation, 2-meter air temperature (Ta), vapor pressure deficit (VPD), incoming short-wave radiation from ERA5 reanalysis data~\cite{hersbach2020era5}, and Leaf Area Index (LAI) data from MODIS~\cite{mkp15}. We extract the time series data closest to each tower site, following a similar approach to~\cite{nathaniel2023metaflux}.

Out of the 206 flux towers in the FLUXNET2015 dataset, we selected 150 towers with data spanning more than three years. For analysis, we considered the most recent three years of data for each selected flux tower. These towers were divided into training and test sets, following guidelines outlined in~\cite{nathaniel2023metaflux}. This process involved selecting half of the towers from tropical and semi-arid regions, which are typically sparse, and one tower from each plant functional type (PFT), including those within cropland and boreal areas, for testing (15 sites). The remaining towers constituted the training set (145 sites). Additionally, we divided the test site data into two blocks: a two-year block representing different scenarios of "few-shot" data availability for testing purposes and one one-year block held out set for testing.

To evaluate our machine learning models on the held-out test year on test sites, we utilized limited years of data (ranging from one month to 24 months) for model adaptation from the "few-shot" data, as illustrated in Figure~\ref{fig:experiment setting}.

\subsection{CARAVAN-GB}
CARAVAN is a recent global benchmark dataset comprising 6,830 basins from various open datasets worldwide, including regions such as the US, UK, GB, Australia, Brazil, and Chile. This dataset provides meteorological forcing data (e.g., precipitation, potential evaporation, temperature), streamflow observations, and basin characteristics. For our study, we focused on a subset consisting of 408 basins from the UK within CARAVAN, referred to as CARAVAN-GB, in our subsequent discussions.
In this dataset, In this dataset, our goal was to estimate streamflow on a daily timescale for the CARAVAN GB Basin. Table ~\ref{tab:forcing_data} shows the meteorological forcing data used in our real-world hydrology experiment with CaravanGB. Our study analyzed data spanning from October 1st, 1989, to September 30th, 2009. The model training phase employed data from 1989 to 1999, while the testing phase encompassed the years 1999 to 2009. The training period from 1989 to 1999 was further divided into training years (1989-1997) and validation years (1997-1999). Out of the 408 available Basins in CARAVAN-GB, we selected 255 basins for our experiment based on the absence of missing streamflow observations between 1989 and 1999. The experimental setting, depicted in Fig~\ref{fig:experiment setting}, involved dividing the basins into train and test sets. The ML models were trained on train basins during train years. The selection of train and test basins was based on mean streamflow, with a 4:1 ratio resulting in 204 basins for training and 51 basins for testing (test sites). We employed a k-fold approach (k=5) to evaluate all models' performance across all basins. For model performance evaluation, the test set employed limited years of data (one year, two years, five years) from the training period, as depicted in Fig~\ref{fig:experiment setting}, for fine-tuning and inferring characteristics as needed within the model. We evaluate on out-of-sample testing blocks as shown in experimental setting Fig~\ref{fig:experiment setting}. 
\begin{table}[h]
    \resizebox{\linewidth}{!}{
        \begin{tabular}{|l|l|}
        \hline
        \textbf{Meteorological forcing data}                                               & \textbf{Unit} \\ \hline
        Daily Precipitation sum(total\_precipitation\_sum)                                 & $mm/day$        \\ \hline
        Daily Potential evaporation sum(potential\_evaporation\_sum)                       & $mm/day$        \\ \hline
        Mean Air temperature (temperature\_2m\_mean)                                       & $^\circ C$            \\ \hline
        Mean Dew point temperature (dewpoint\_temperature\_2m\_mean)                       & $^\circ C$            \\ \hline
        Mean Shortwave radiation (surface\_net\_solar\_radiation\_mean)                    & $Wm^{-2}$          \\ \hline
        Mean Net thermal radiation at the surface (surface\_net\_thermal\_radiation\_mean) & $Wm^{-2}$          \\ \hline
        Mean Surface pressure (surface\_pressure\_mean)                                    & $kPa$           \\ \hline
        Mean Eastward wind component (u\_component\_of\_wind\_10m\_mean)                   & $ms^{-1}$          \\ \hline
        Mean Northward wind component (v\_component\_of\_wind\_10m\_mean)                  & $ms^{-1}$         \\ \hline
        \end{tabular}
    }
    \caption{\small A table of meteorological forcing data used in this experiment.}
    \label{tab:forcing_data}
    \vspace{-0.2cm}
\end{table}

\subsection{Synthetic Dataset:} We follow the synthetic data generation setup used by Vuorio et al.~\cite{vuorio2019multimodal}. Specifically, we create a set of task distributions for regression with different modes. Modes are a collection of functions that include sinusoidal functions, linear functions, quadratic functions, $\ell1$ norm functions, and hyperbolic tangent functions. Each task/function from these modes is created by changing as shown:\\
\noindent\textbf{Sinusoidal:} $A \cdot \sin{w \cdot x + b} + \epsilon$, with $A \in [0.1, 5.0]$, $w \in [0.5, 2.0]$ and $b \in [0, 2\pi ]$\\
\noindent\textbf{Linear:} $A \cdot x + b$, with $A \in [-3, 3]$\\
\noindent\textbf{Quadratic:} $A \cdot (x - c)^2 + b$, with $A \in [-0.15, -0.02] \cup [0.02, 0.15]$, $c \in [-3.0, 3.0]$ and $b \in [-3.0, 3.0]$ )\\
\noindent\textbf{$\ell1$ norm:} $A \cdot |x - c| + b$, with $A \in [-0.15, -0.02] \cup [0.02, 0.15]$, $c \in [-3.0, 3.0]$ and $b \in [-3.0, 3.0]$\\
\noindent\textbf{Hyperbolic tangent:} $A \cdot tanh(x - c) + b$, with $A \in [-3.0, 3.0]$, $c \in [-3.0, 3.0]$ and $b \in [-3.0, 3.0]$\\

\vspace{-0.3cm}
We add Gaussian noise ($\mu$ = 0 and standard deviation = 0.3) to each data point obtained from the individual task.

\section{Experimental setting}
\label{sec:experimental_setting}
\subsection{Hyperparameter Tuning}

\subsubsection{Experiment: Flux Tower Dataset}

\begin{table}[h]
\scriptsize
\resizebox{\linewidth}{!}{
\begin{tabular}{|l|l|}
\hline
\textbf{Hyperparameter}                     & \textbf{Value}             \\ \hline
Dimension of Encoder                 & 16, 32, \textbf{64}                 \\ \hline
Dimension of hidden state for Forward Model & 128, 256, \textbf{350}, 500               \\ \hline
Batch size                                  & 32, \textbf{64}, 128                \\ \hline
Learning rate                               & 0.005, \textbf{0.001}, 0.0005, 0.01 \\ \hline
\end{tabular}
}
\caption{\small Range of parameter values tried for hyperparameter tuning, with the final selected value shown in \textbf{bold}.}
\vspace{-0.2cm}
\label{tab:hyperparameter_tuning_flux}
\end{table}

To find the best hyperparameters, we performed a grid search across a range of parameter values. The possible values considered are listed in Table~\ref{tab:hyperparameter_tuning_flux}. We trained our model, TAM-RL, using data from training sites. The final hyperparameters, including batch size and learning rates, are determined after performing a k-fold cross-validation (k=5)
on the training set. Finally, we evaluated the model's performance on the testing sites during the hold-out test period, as shown in Figure~\ref{fig:experiment setting}.\\
\subsubsection{Experiment: CARAVAN GB}

\begin{table}[h]
\scriptsize
\resizebox{\linewidth}{!}{
\begin{tabular}{|l|l|}
\hline
\textbf{Hyperparameter}                     & \textbf{Value}             \\ \hline
Latent dimension of Encoder                 & 16, \textbf{32}, 64                 \\ \hline
Dimension of hidden state for Forward Model & 64, \textbf{128}, 256               \\ \hline
Batch size                                  & 32, \textbf{64}, 128                \\ \hline
Learning rate                               & 0.005, \textbf{0.001}, 0.0005, 0.01 \\ \hline
\end{tabular}
}
\caption{\small Range of parameter values tried for hyperparameter tuning, with the final selected value shown in \textbf{bold}.}
\vspace{-0.2cm}
\label{tab:hyperparameter_tuning_caravan}
\end{table}

To find the best hyperparameters, we performed a grid search across a range of parameter values. The possible values considered are listed in Table~\ref{tab:hyperparameter_tuning_caravan}. We trained our model, TAM-RL, using data from the training basins during the training period. The final hyperparameter configuration was chosen based on the set with the lowest average RMSE in the validation period for the training basins. Finally, we evaluated the model's performance on the testing basins during the test period, as shown in Figure~\ref{fig:experiment setting}.\\

\subsubsection{Experiment: Synthetic Dataset}
We use the same hyperparameter settings as
the regression experiments presented in ~\cite{vuorio2019multimodal} and used Adam ~\cite{kingma2014adam} as the meta-optimizer. 

\subsection{Network Architectures}
\subsubsection{Base Network}
For the Flux Tower Dataset, we use an LSTM as the base network with a hidden dimension of 350; for the Caravan-GB dataset, we use an LSTM as the base network with a hidden dimension of 128.
For the synthetic dataset, Like in the case of ~\cite{vuorio2019multimodal}, we use a 4-layer fully connected neural network with a hidden dimension of 100 and ReLU non-linearity for each layer as a base network.

\subsubsection{Task Encoder}
We use a BiLSTM as the base network for all three datasets, where Flux towers BiLSTM uses 64 hidden dimensions,  Caravan-GB uses a BiLSTM of 32 hidden dimensions and the synthetic dataset uses BiLSTM of hidden dimension 40.

\subsubsection{Modulation}
For few-shot learning, it is essential to modulate the base network in a way that is parameter-efficient. Unconstrained modulation, while being flexible, can also be slow and prone to overfitting. Instead, we propose using Film modulation~\cite{perez2018film} of the LSTM gate outputs for LSTM-based base networks, and which requires only a small number of task-specific parameters for both real-world problems. We similarly also perform Film modulation at each layer for the fully connected neural network-based base network.

\subsubsection{Parameter Generator}
All models have an MLP layer as a parameter generator. Since modulation occurs in two stages, according to our modulation strategy, we employ two fully connected linear layers to convert the task embedding vector z to a vector of hidden dimension 700, i.e., 2*(64,700). The hidden dimension of the LSTM base network was 350, and for film modulation, we need twice the hidden dimension. In our implementation, we use half the dimension of the vector as two components for FILM modulation. 
Similarly, the Hydrology dataset also uses an LSTM as a base network. So, we use two linear, fully connected layers of dim (32,256).
The synthetic dataset has four convolutional layers with channel sizes 32, 64, 128, and 256. To modulate them, we create four linear layers with output dimensions 64, 128, 256, and 512, respectively.

\section{Pseudocode}
\subsection{Training Algorithm}
Algorithm~\ref{alg:training_pseudo} shows the training procedure followed in TAM-RL.

\begin{algorithm}[ht]
    \caption{TAM-RL Training Algorithm with Pretraining}\label{alg:cap}
    \small
    \begin{algorithmic}
        \Require $D^{Train}$: Full training dataset, $W^{Tr_{\mathcal{T}}}=[W^{support_{\mathcal{T}}},W^{query_{\mathcal{T}}}]$ generated from $D^{Train}$, $\alpha$: step size hyper-parameter; We abbreviate support as s and query as q
        
        \State // Pretraining phase
        \State Randomly initialize base network weights $\theta$ (LSTM-based)
        \For {epoch = 1 to $N_{pretrain}$}
            \State Sample batch of data $(X, Y)$ from $D^{Train}$
            \State Update $\theta$ with $\alpha \nabla_{\theta} \mathcal{L}(\mathcal{F}(X; \theta), Y)$
        \EndFor
        
        \State // Joint training phase
        \State Randomly initialize $\phi_{\tau}$ and $\phi_z$
        \For {epoch = 1 to $N_{joint}$}
            \While{not Done}
                \State Sample batches of entities $W_k^{Tr_{\mathcal{T}}} \sim W^{Train_{\mathcal{T}}}$
                \For {all k}
                    \State Infer $\boldsymbol{z} = \mathcal{E}(W_k^{s_{\mathcal{T}}};\phi_{z})$
                    \State Infer $\tau = \mathcal{G}(\boldsymbol{z};\phi_{\tau})$
                    \State Compute $\theta'_k = \theta \bigotimes \tau$ (modulation)
                    \State $\hat{y}_k^{q_{\mathcal{T}}} = \mathcal{F}(x_k^{q_{\mathcal{T}}}; \theta'_k)$
                    \State Update $\theta$, $\phi_{\tau}$, and $\phi_h$ with $\alpha \nabla_{\theta,\phi_{\tau},\phi_z} \mathcal{L}_{W_k^{Tr_{\mathcal{T}}}}(\hat{y}_k^{q_{\mathcal{T}}};W_k^{q_{\mathcal{T}}})$
                \EndFor
            \EndWhile
        \EndFor
    \end{algorithmic}
    \label{alg:training_pseudo}
\end{algorithm}

\subsection{Inference Algorithm}
Algorithm~\ref{alg:inference} shows the inference procedure followed in TAM-RL.

\begin{algorithm}
    \caption{TAM-RL Inference Algorithm}\label{alg:cap}
    \small
    \begin{algorithmic}
        \Require $W_j^{{support}_{\mathcal{T}}}$ generated from $D_j^{Few}$, $x_j^{q_{\mathcal{T}}}$, $\beta$: step size hyper-parameter, num\_inner\_steps: no. of gradient steps for the inner loop
        \Ensure Output $y_j^{q_{\mathcal{T}}}$ for input driver $x_j^{q_{\mathcal{T}}}$ of an entity j
        \State Load $\theta$, $\phi_{\tau}$, and $\phi_z$ from trained model
        \State Infer $\boldsymbol{z} = \mathcal{E}(W_j^{s_{\mathcal{T}}};\phi_{z})$
        \State Infer $\tau = \mathcal{G}(\boldsymbol{z};\phi_{\tau})$
        \State Compute $\theta'_j = \theta \bigotimes \tau$ (modulation)
        \For {step = 1 to num\_inner\_steps}
            \State Update $\theta'_j$ with $\beta \nabla_{\theta'_j} \mathcal{L}_{W_j^{s_{\mathcal{T}}}}(\mathcal{F}(x_j^{s_{\mathcal{T}}} ; \theta'_j);W_j^{s_{\mathcal{T}}})$
        \EndFor
        \State \Return $y_j^{q_{\mathcal{T}}} = \mathcal{F}(x_j^{q_{\mathcal{T}}} ; \theta'_j)$
    \end{algorithmic}
    \label{alg:inference}
\end{algorithm}

\input{tables/few_shot_synthetic_expanded}
\section{Few-shot Evaluation on Synthetic Dataset} 
\begin{figure}
    \centering
    \includegraphics[width=0.9\linewidth]{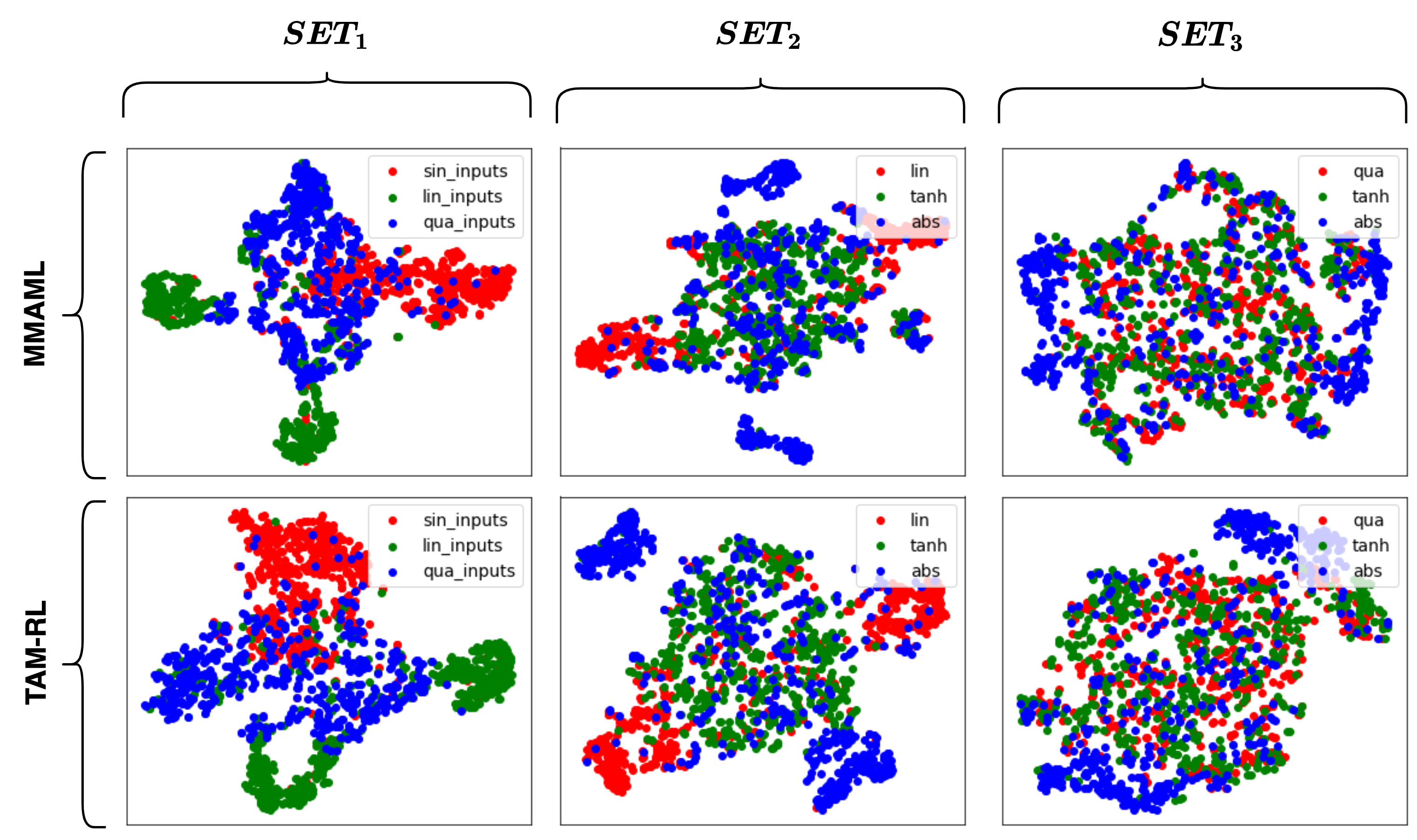}
    \caption{\footnotesize \textit{tSNE plots of the task embeddings produced by MMAML (top row) and TAM-RL (bottom row) for the three sets of multimodal tasks (shown in columns).}}
    \label{fig:tsne_l4}
\end{figure}
The design choice of TAM-RL, which leverages meta-initializations directly through an amortization approach, proves highly effective in generating superior meta-initializations for diverse tasks through learned distinct representations. To rigorously evaluate the effectiveness of TAM-RL, we designed a controlled synthetic data experiment by systematically varying the degree of task heterogeneity. We train MAML, MMAML, and TAM-RL for each of the three sets of multimodal distributions, SET$_1$, SET$_2$, and SET$_3$ mentioned in Section~\ref{sec:experiments}. Following~\cite{vuorio2019multimodal}, we use an MLP-based base network for all three models: MAML, MMAML, and TAM-RL. For MMAML and TAM-RL, Bidirectional LSTM is used as the task encoder (Please refer to the appendix for details of the architecture). We use ADAM as the meta-optimizer and use the same hyperparameter settings as the regression experiments in ~\cite{vuorio2019multimodal}. We evaluate all models with five gradient steps during fine-tuning using five meta-train examples for these new tasks. As the evaluation criterion, we report the mean squared error (MSE) of five meta-validation examples of these new tasks.

Fig~\ref{fig:tsne_l4} shows the tSNE plots~\cite{van2008visualizing} of the task embeddings produced by TAM-RL and MMAML, respectively, from randomly sampled tasks for each set. From the tSNE plot, we observe that as we move from $SET_1$ to $SET_3$, the class of functions becomes more and more homogeneous. For example, a quadratic function can resemble a sinusoidal or linear function, but generally, a sinusoidal function is dissimilar from a linear function. $SET_3$ is homogenous because quadratic, $\ell1$, and hyperbolic tangent functions are very similar, particularly with Gaussian noise in the output space. Further, we can observe from the tSNE plots that TAM-RL and MMAML task embeddings generate similar tSNE plots and clusters.

We report the mean MSE for each architecture on all three sets in Table~\ref{tab:few-shot_synthetic_evaluation}. We can observe that $SET_1$ has a reasonably heterogeneous split, as shown in Fig~\ref{fig:tsne_l4}. For SET$_1$, the performance of TAM-RL is 21.6\% better than MMAML. As the set becomes more homogeneous, like in SET$_2$, the performance gap between TAM-RL and MMAML narrows and, in this scenario, TAM-RL only holds a slight edge of 3.5\% over MMAML, and they operate at a similar performance level. When the dataset reaches near complete homogeneity, as in SET$_3$, TAM-RL achieves similar performance to MMAML, with TAM-RL being worse by 5.9\%. From Table~\ref{tab:few-shot_synthetic_evaluation}, we can also observe that MAML has the highest error in all settings and that incorporating task identity through task-specific parameters matters significantly in multi-modal task distributions. Further, our results show that TAM-RL outperforms MMAML as the set becomes more heterogeneous.

Here, we also show Table ~\ref{tab:expanded_sythetic_table}, an expanded version of Table ~\ref{tab:few-shot_synthetic_evaluation} with task family-wise MSEs.

\section{Streamflow Prediction Plot}
\begin{figure*}
        \centering
    \begin{subfigure}{0.8\linewidth}
        \centering
        \includegraphics[width=\linewidth]{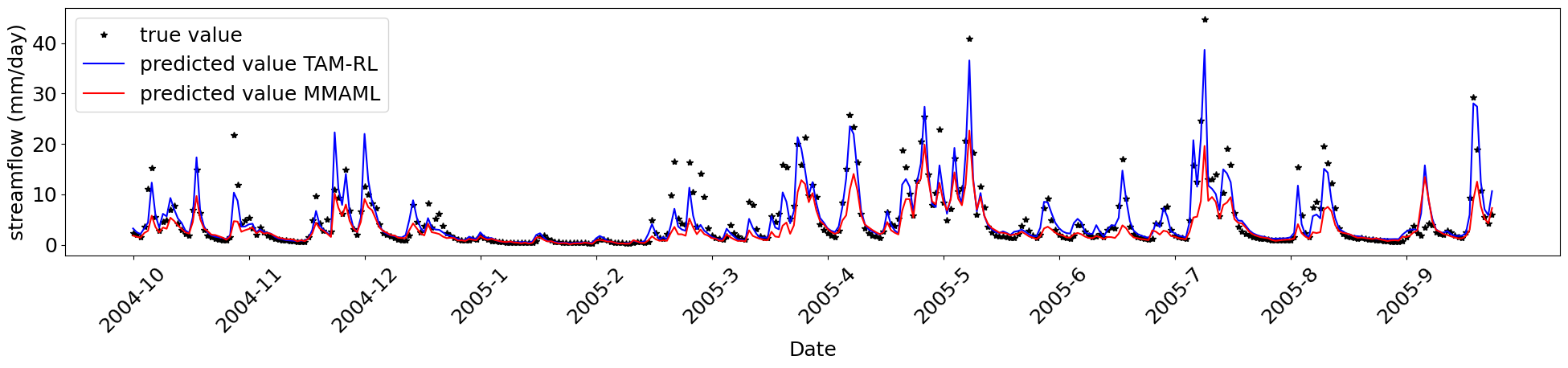}
        \caption{}
        \label{fig:base}
    \end{subfigure}    
    \begin{subfigure}{0.8\linewidth}
        \centering
        \includegraphics[width=\linewidth]{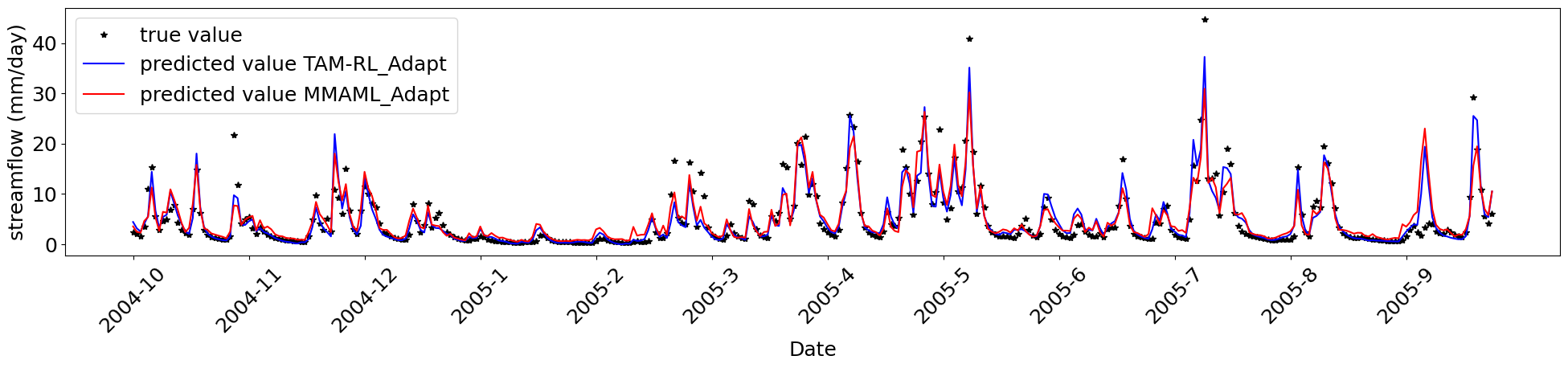}
        \caption{}
        \label{fig:adapt}    
    \end{subfigure}
    \caption{\small Observed Streamflow and predicting streamflow by different model architectures a) Models w/o fine-tuning b)Models with fine-tuning (Best seen in color)}
    \vspace{-0.5cm}
    \label{fig:streamflow}
\end{figure*}

Figure~\ref{fig:base} shows the prediction from the models without fine-tuning. We can observe that the TAM-RL predictions match the observed streamflow better than MMAML. Figure~\ref{fig:adapt} shows the prediction from the models on adapting with only two years of data. From the plot, we observe that all models' prediction improves in general. However, TAM-RL gives a prediction closer to the ground truth. 

\vspace{-0.2cm}
\section{Reproducibility}
CARAVAN dataset is freely available at zenodo at \footnote{\url{https://doi.org/10.5281/zenodo.7540792}}.
The code used in this paper is available at Google Drive \footnote{\url{https://drive.google.com/drive/folders/1Z\_Uuq6qEzCkaiPYAtPapgEiIZzZZnXxy}}.

%% file: tables/few_shot_synthetic_expanded.tex
\begin{table*}[t!]
\resizebox{\textwidth}{!}{%
\begin{tabular}{|c|ccc|ccc|ccc|}
\hline
\multirow{2}{*}{Architecture} & \multicolumn{3}{c|}{\textbf{$\boldsymbol {SET_1}$ \{Sin, Linear, Quadratic\}}}                                             & \multicolumn{3}{c|}{\textbf{$\boldsymbol{SET_2}$ \{Linear,  Tanh,$\ell1$ norm\}}}                                                    & \multicolumn{3}{c|}{\textbf{$\boldsymbol{SET_3}$ \{Quadratic, Tanh, $\ell1$ norm\}}}                                      \\ \cline{2-10} 
                              & \multicolumn{1}{c|}{\textbf{Sin}}         & \multicolumn{1}{c|}{\textbf{Linear}}      & \textbf{Quadratic}  & \multicolumn{1}{c|}{\textbf{Linear}}      & \multicolumn{1}{c|}{\textbf{Tanh}}       & \textbf{$\ell1$ norm}             & \multicolumn{1}{c|}{\textbf{Quadratic}}  & \multicolumn{1}{c|}{\textbf{Tanh}}        & \textbf{$\ell1$ norm} \\ \hline
\multirow{2}{*}{$MAML$}   & \multicolumn{1}{c|}{5.623 $\pm$ 2.40e-01} & \multicolumn{1}{c|}{4.880 $\pm$ 4.81e-01} & 0.550 $\pm$ 4.6e-02 & \multicolumn{1}{c|}{7.216 $\pm$ 3.39e-01} & \multicolumn{1}{c|}{1.489 $\pm$ 5.2e-02} & 1.63 \$\textbackslash{}pm 8.5e-02 & \multicolumn{1}{c|}{0.913 $\pm$ 2.5e-02} & \multicolumn{1}{c|}{1.484 $\pm$ 4.3e-02}  & 1.965 $\pm$ 1.11e-01  \\ \cline{2-10} 
                              & \multicolumn{3}{c|}{Set Mean MSE: 3.684 $\pm$ 1.76e-01}                                                     & \multicolumn{3}{c|}{Set Mean MSE: 3.446 $\pm$ 1.15e-01}                                                                  & \multicolumn{3}{c|}{Set Mean MSE: 1.454 $\pm$ 4.0e-02}                                                       \\ \hline
\multirow{2}{*}{$MMAML$}  & \multicolumn{1}{c|}{0.815 $\pm$ 2.08e-01} & \multicolumn{1}{c|}{0.440 $\pm$ 6.9e-02}  & 0.547 $\pm$ 4.5e-02 & \multicolumn{1}{c|}{0.664 $\pm$ 7.8e-02}  & \multicolumn{1}{c|}{0.697 $\pm$ 6.9e-02} & 1.895 $\pm$ 3.55e-01              & \multicolumn{1}{c|}{0.581 $\pm$ 5.4e-02} & \multicolumn{1}{c|}{0.977 $\pm$ 1.02e-01} & 0.713 $\pm$ 9.3e-02   \\ \cline{2-10} 
                              & \multicolumn{3}{c|}{Set Mean MSE: 0.601 $\pm$ 7.4e-02}                                                      & \multicolumn{3}{c|}{Set Mean MSE: 1.085 $\pm$ 1.21e-01}                                                                  & \multicolumn{3}{c|}{Set Mean MSE: \textbf{0.757 $\boldsymbol{\pm}$  5.0e-02}}                                                       \\ \hline
\multirow{2}{*}{$TAM-RL$ (ours)}       & \multicolumn{1}{c|}{0.553 $\pm$ 1.00e-01} & \multicolumn{1}{c|}{0.391 $\pm$ 4.6e-02}  & 0.539 $\pm$ 4.5e-02 & \multicolumn{1}{c|}{0.873 $\pm$ 9.3e-02}  & \multicolumn{1}{c|}{0.577 $\pm$ 3.2e-02} & 1.696 $\pm$ 1.84e-01              & \multicolumn{1}{c|}{0.546 $\pm$ 2.3e-02} & \multicolumn{1}{c|}{1.051 $\pm$ 4.7e-02} & 0.817 $\pm$ 4.5e-02   \\ \cline{2-10} 
                              & \multicolumn{3}{c|}{Set Mean MSE: \textbf{0.494 $\boldsymbol{\pm}$ 4.0e-02}}                                                      & \multicolumn{3}{c|}{Set Mean MSE: \textbf{1.049 $\boldsymbol{\pm}$  6.9e-01}}                                                                   & \multicolumn{3}{c|}{Set Mean MSE: 0.805 $\pm$ 2.3e-02}                                                       \\ \hline
\end{tabular}%
}
\caption{Mean square error (MSE) on the multimodal 5-shot regression with different combinations of 3 modes for different architectures. Gaussian noise is applied to each function with $\mu=0$ and $\sigma=0.3$.}
\label{tab:expanded_sythetic_table}
\end{table*}